\def\year{2017}\relax

\documentclass[letterpaper]{article}
\usepackage{aaai17}
\usepackage{times}
\usepackage{helvet}
\usepackage{courier}
\usepackage{graphicx}
\usepackage{multirow}
\usepackage{float}
\usepackage{amsmath}
\usepackage{color}

\usepackage{hyperref}

\newcommand{\layername}[1]{\texttt{#1}}

\newcommand{\smallcite}[1]{{\tiny \cite{#1}}}

\begin{document}
%
\title{TextBoxes: A Fast Text Detector with a Single Deep Neural Network}
\author{Minghui Liao\thanks{Authors contribute equally.}, Baoguang Shi\footnotemark[\value{footnote}], Xiang Bai\thanks{Corresponding author.}, Xinggang Wang, Wenyu Liu\\
School of Electronic Information and Communications, Huazhong University of Science and Technology\\
\{mhliao, xbai, xgwang, liuwy\}@hust.edu.cn \and shibaoguang@gmail.com
}

\maketitle

\begin{abstract}
This paper presents an end-to-end trainable fast scene text detector, named TextBoxes, which detects scene text with both high accuracy and efficiency in a single network forward pass, involving no post-process except for a standard non-maximum suppression. TextBoxes outperforms competing methods in terms of text localization accuracy and is much faster, taking only 0.09s per image in a fast implementation. Furthermore, combined with a text recognizer, TextBoxes significantly outperforms state-of-the-art approaches on word spotting and end-to-end text recognition tasks.
\end{abstract}

\section{Introduction}

Scene text is one of the most general visual objects in natural scenes. It frequently appears on road signs, license plates, product packages, \emph{etc}. Reading scene text facilitates a lot of useful applications, such as image-based geolocation. Despite the similarity to traditional OCR, scene text reading is much more challenging, due to the large variations in both foreground text and background objects, as well as uncontrollable lighting conditions, \emph{etc}.

Owing to the inevitable challenges and complexities, traditional text detection methods tend to involve multiple processing steps, \emph{e.g.} character/word candidate generation~\cite{neumann2012real,jaderberg2016reading}, candidate filtering, and grouping. They often end up struggling to get each module working properly, requiring much effort in tuning parameters and designing heuristic rules, also slowing down detection speed. Inspired by the recent developments in object detection~\cite{liu2015ssd,ren2015faster}, we propose to detect texts by \emph{directly} predicting word bounding boxes via a single neural network that is end-to-end trainable.

Our key contribution in this paper is a fast and accurate text detector called \emph{TextBoxes}, which is based on fully-convolutional network~\cite{lecun1998gradient}. TextBoxes directly outputs the coordinates of word bounding boxes at multiple network layers by jointly predicting text presence and coordinate offsets to \emph{default boxes}~\cite{liu2015ssd}. The final outputs are the aggregation of all boxes, followed by a standard non-maximum suppression process. To handle the large variation in aspect ratios of words, we design several novel, inception-style~\cite{szegedy2015going} output layers that utilize both irregular convolutional kernels and default boxes. Our detector delivers both high accuracy and high efficiency with only a single forward pass on single-scale inputs, and even higher accuracy with multiple passes on multi-scale inputs.

Furthermore, we argue that word recognition is helpful to distinguish texts from backgrounds, especially when words are confined to a given set, \emph{i.e.} a lexicon. We adopt a successful text recognition algorithm, CRNN~\cite{shi2015end}, in conjunction with TextBoxes. The recognizer not only provides extra recognition outputs, but also regularizes text detection with its semantic-level awareness, thus further boosting the accuracy of word spotting considerably. The combination of TextBoxes and CRNN yields the state-of-the-art performance on word spotting and end-to-end text recognition tasks, which appears to be a simple yet effective solution to robust text reading in the wild.

To summarize, the contributions of this paper are three-fold: First, we design an end-to-end trainable neural network model for scene text detection. Second, we propose a word spotting/end-to-end recognition framework that effectively combines detection and recognition. Third, our model achieves highly competitive results while keeping its computational efficiency.

\begin{figure*}[!ht]
\begin{center}
\includegraphics[width=0.7\linewidth]{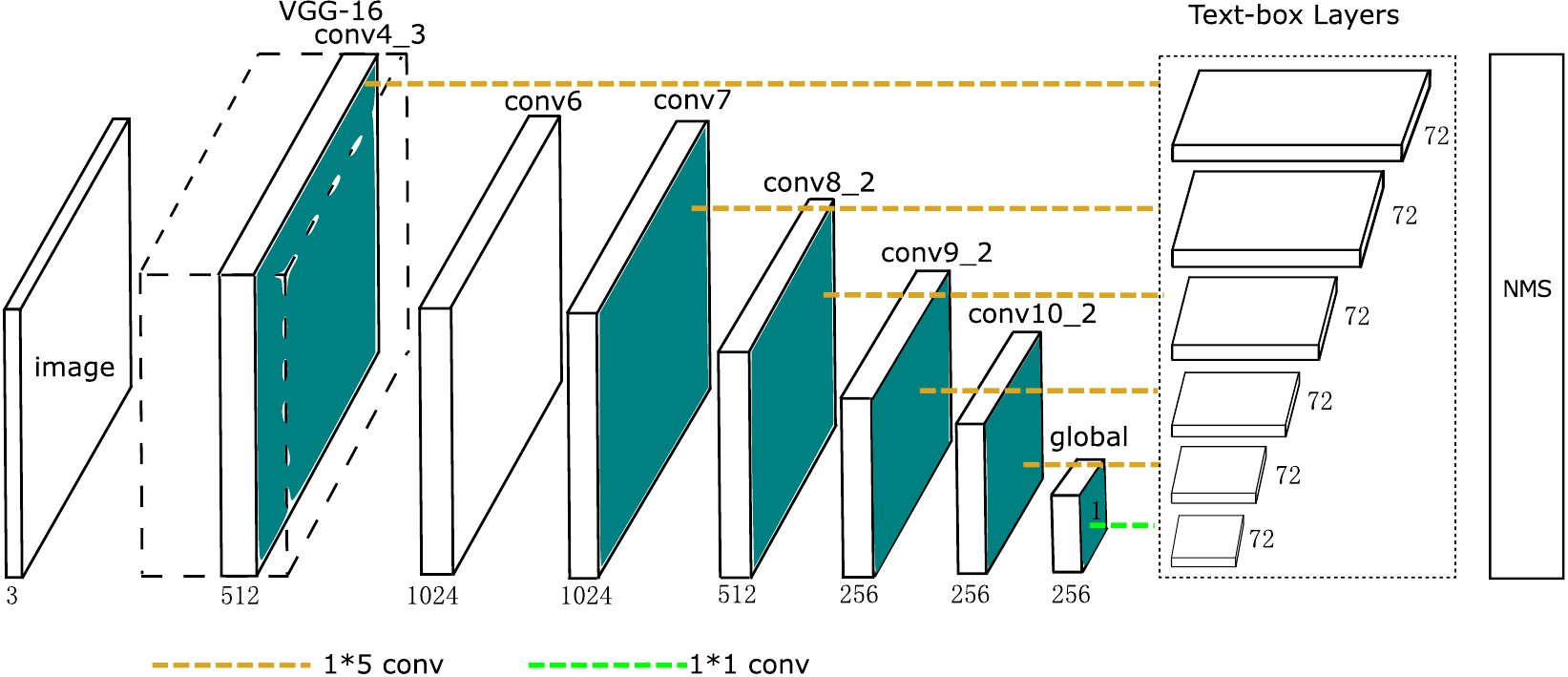}
\end{center}
\vspace{-5mm}
\caption{TextBoxes Architecture. TextBoxes is a 28-layer fully convolutional network. Among them, 13 are inherited from VGG-16. 9 extra convolutional layers are appended after the VGG-16 layers. Text-box layers are connected to 6 of the convolutional layers. On every map location, a text-box layer predicts a 72-d vector, which are the text presence scores (2-d) and offsets (4-d) for 12 default boxes. A non-maximum suppression is applied to the aggregated outputs of all text-box layers.
}
\label{fig:pipeline}
\vspace{-2mm}
\end{figure*}

\section{Related Works}

Intuitively, scene text reading can be further divided into two sub-tasks: text detection and text recognition. The former aims to localize text in images, mostly in the form of word bounding boxes; The latter transcripts cropped word images into machine-interpretable character sequences. We cover both tasks in this paper but pay more attention to detection.

Based on a basic detection target, previous methods for text detection can be roughly categorized into three categories:\\
\emph{1) Character-based}: Individual characters are first detected and then grouped into words~\cite{neumann2012real,Pan2011,Yao2012,huang2014robust}. For example, \cite{neumann2012real} locates characters by classifying Extremal Regions. After that, the detected characters are grouped by an exhaustive search method;\\
\emph{2) Word-based}: Words are directly hit with the similar manner of general object detection \cite{jaderberg2016reading,Zhong2016,TextProposal}. \cite{jaderberg2016reading} proposes an R-CNN-based~\cite{girshick14CVPR} framework.First, word candidates are generated with class-agnostic proposal generators. Then the proposals are classified by a random forest classifier. Finally, a convolutional neural network for bounding box regression was adopted to refine the bounding boxes. \cite{gupta2016synthetic} improves over the YOLO network~\cite{RedmonDGF15} while it still adopts the filter and regression steps for further removing the false positives;\\
\emph{3) Text-line-based}: Text lines are detected and then broken into words. For example,~\cite{zhang2015symmetry} proposes to detect text lines utilizing their symmetric characteristics. Furthermore, \cite{Zhang_2016_CVPR} localizes text lines with fully convolutional networks~\cite{long2015fully}.

TextBoxes is \emph{word-based}. In contrast to~\cite{jaderberg2016reading}, which comprises three detection steps and each further includes more than one algorithm, TextBoxes enjoys a much simpler pipeline. We only need to train one network end-to-end.

TextBoxes is inspired by SSD~\cite{liu2015ssd}, a recent development in object detection. SSD aims to detect general objects in images but fails on words that have extreme aspect ratios. We propose text-box layers in TextBoxes to solve this problem, which significantly improve the performance.

We adopt a text recognizer called CRNN~\cite{shi2015end} in conjunction with TextBoxes for word spotting and end-to-end recognition. CRNN directly outputs character sequences given input images and is also end-to-end trainable. Besides, we use the confidence scores of CRNN to regularize the detection outputs of TextBoxes. Note that it is also possible to adopt other recognizers, such as~\cite{jaderberg2016reading}.

\section{Detecting text with TextBoxes}

\subsection{Architecture}

The architecture of TextBoxes is depicted in Fig.~\ref{fig:pipeline}. It inherits the popular VGG-16 architecture~\cite{simonyan2014very}, keeping the layers from \layername{conv1\_1} through \layername{conv4\_3}. The last two fully-connected layers of VGG-16 are converted into convolutional layers by parameters down-sampling~\cite{liu2015ssd}. They are followed by a few extra convolutional and pooling layers, namely \layername{conv6} to \layername{pool11}.

Multiple output layers, which we call \emph{text-box layers}, are inserted after the last and some intermediate convolutional layers. Their outputs are aggregated and undergo a non-maximum suppression (NMS) process. Output layers are also convolutional. All together, TextBoxes consists of only convolutional and pooling layers, thus \emph{fully-convolutional}. It adapts to arbitrary-size images in both training and testing.

\subsection{Text-box layers}

Text-box layers are the key component of TextBoxes. A text-box layer simultaneously predicts text presence and bounding boxes, conditioned on its input feature map.
At every map location, it outputs the classification scores and offsets to its associated default boxes in a convolutional manner.
Suppose that image and feature map sizes are respectively $(w_\textrm{im}, h_\textrm{im})$ and $(w_\textrm{map}, h_\textrm{map})$. On a map location $(i,j)$ which associates a default box $\mathbf{b}_0=(x_0, y_0, w_0, h_0)$, the text-box layer predicts the values of $(\Delta x, \Delta y, \Delta w, \Delta h, c)$, indicating that a box $\mathbf{b}=(x, y, w, h)$ is detected with confidence $c$, where

\begin{align}
\begin{split}
  x &= x_0 + w_0 \Delta x, \\
  y &= y_0 + h_0 \Delta y, \\
  w &= w_0 \exp(\Delta w), \\
  h &= h_0 \exp(\Delta h).
\end{split}
\label{eq:decode-box}
\end{align}

In the training phase, ground-truth word boxes are matched to default boxes according to box overlap, following the matching scheme in~\cite{liu2015ssd}. Each map location is associated with multiple default boxes of different sizes. They effectively divide words by their scales and aspect ratios, allowing TextBoxes to learn specific regression and classification weights that handle words of similar size. Therefore, the design of default boxes is highly task-specific.

\begin{figure}[!htbp]
\begin{center}
\includegraphics[width=0.6\linewidth]{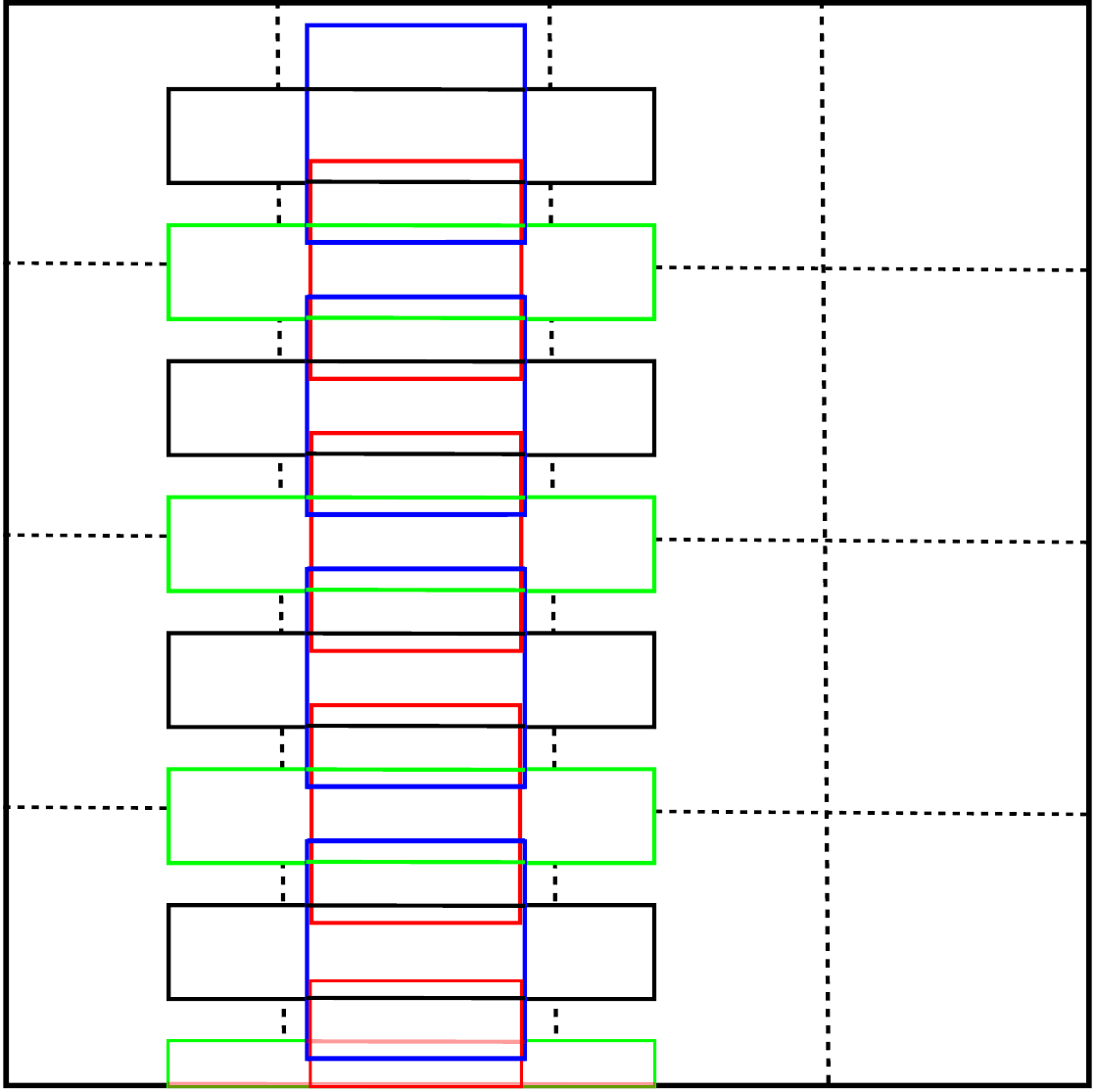}
\end{center}
\vspace{-5mm}
\caption{Illustration of default boxes for a 4*4 grid.
For better visualization, only a column of default boxes whose aspect ratios 1 and 5 are plotted. The rest of the aspect ratios are 2,3,7 and 10, which are placed similarly.
The black (aspect ratio: 5) and blue (ar: 1) default boxes are centered in their cells. The green (ar: 5) and red (ar: 1) boxes have the same aspect ratios and a vertical offset(half of the height of the cell) to the grid center respectively.
}
\label{fig:default-boxes}
\vspace{-2mm}
\end{figure}

Different from general objects, words tend to have large aspect ratios. Therefore, we include ``long'' default boxes that have large aspect ratios. Specifically, we define 6 aspect ratios for default boxes, including 1,2,3,5,7, and 10. However, this makes the default boxes dense on the horizontal direction while sparse vertically, which causes poor matching boxes. To solve this issue, each default box is set with vertical offsets. The design of the default boxes is illustrated in Fig.~\ref{fig:default-boxes}.

Moreover, in text-box layers we adopt irregular 1*5 convolutional filters instead of the standard 3*3 ones. This inception-style~\cite{szegedy2015going} filters yield rectangular receptive fields, which better fit words with larger aspect ratios, also avoiding noisy signals that a square-shaped receptive field would bring in.

\subsection{Learning}
We adopt the same loss function as~\cite{liu2015ssd}. Let $x$ be the match indication matrix, $c$ be the confidence, $l$ be the predicted location, and $g$ be the ground-truth location. Specifically, for the $i$-th default box and the $j$-th ground truth, $x_{ij}=1$ means matching while $x_{ij}=0$ otherwise. The loss function is defined as:

\begin{equation}
L(x,c,l,g)=\frac{1}{N}(L_{\textrm{conf}}(x,c)+\alpha L_{\textrm{loc}}(x,l,g)),
\end{equation}
where $N$ is the number of default boxes that match ground-truth boxes, and $\alpha$ is set to 1. We adopt the smooth L1 loss~\cite{fast_rcnn} for $L_{\textrm{loc}}$ and a 2-class softmax loss for $L_{\textrm{conf}}$.

\subsection{Multi-scale inputs}

Even with the optimizations on default boxes and convolutional filters,  it may be still difficult to robustly localize the words of extreme aspect ratios and sizes. To further boost detection accuracy, we use multiple rescaled versions of the input image for TextBoxes. An input image is rescaled into five scales, including (width*height) 300*300, 700*700, 300*700, 500*700, and 1600*1600. Note that some scales squeeze image horizontally, so that some ``long'' words are shortened. Multi-scale inputs boost detection accuracy while slightly increasing the computational cost. On ICDAR 2013 , they further improve f-measure of detection by 5 percents. Detecting all five scales takes 0.73s per image, and 0.24s if we remove the last 1600*1600 scale. The running time is measured on a single Titan X GPU. Note that, different from testing, we only use single-scale input (300*300) for training. 

\subsection{Non-maximum suppression}
Non-maximum suppression is applied to the aggregated outputs of all text-box layers. We adopt an extra non-maximum suppression for multi-scale inputs on the task of text localization.

\subsection{Word spotting and end-to-end recognition}
Word spotting is to localize specific words that are given in a lexicon. End-to-end recognition concerns both detection and recognition.
Although both tasks can be achieved by simply connecting TextBoxes with a text recognizer,
we propose to improve detection with recognition.
We argue that a recognizer can help eliminating false-positive detection results that are unlikely to be meaningful words, \emph{e.g.} repetitive patterns. Particularly, when a lexicon is present, a recognizer could effectively removes the detected bounding boxes that do not match any of the given words.

We adopt the CRNN model~\cite{shi2015end} as our text recognizer. CRNN uses CTC~\cite{graves2006connectionist} as its output layer, which estimates sequence probability conditioned on input image, \emph{i.e.} $p(\textbf{w} | I)$, where $I$ is an input image and $\textbf{w}$ represents a character sequence. 
We treat the probability as a matching score, which measures the compatibility of an image to a particular word. The detection score is then the maximum score among all words in a given lexicon:
\begin{equation}
  s = \max_{\textbf{w} \in {\cal W} } p(\textbf{w} | I)
\label{eq:ctc-score}
\end{equation}
where ${\cal W}$ is a given lexicon. If the task specifies no lexicon, we use a generic lexicon that consists of 90k English words.

\begin{table*}[!tbp] \small
\centering
\caption{Text localization on ICDAR 2011 and ICDAR 2013. P, R and F refer to precision, recall and F-measure respectively. FCRNall+filts reported a time consumption of 1.27 seconds excluding its regression step so we assume it takes more than 1.27 seconds.}
\label{tab:text-localization}
\renewcommand{\arraystretch}{1.2}
\begin{tabular}{|c|c|c|c|c|c|c|c|c|c|c|c|c|c|}
\hline
\multicolumn{1}{|c|}{Datasets}                                  & \multicolumn{6}{c|}{ICDAR 2011}                                                       & \multicolumn{6}{c|}{ICDAR 2013}                                                       & \multirow{3}{*}{Time/s} \\ \cline{1-13}
Evaluation protocol                                              & \multicolumn{3}{c|}{IC13 Eval}    & \multicolumn{3}{c|}{DetEval}                  & \multicolumn{3}{c|}{IC13 Eval}    & \multicolumn{3}{c|}{DetEval}                  &                         \\ \cline{1-13}
Methods                                                         & P             & R             & F             & P             & R             & F             & P             & R             & F             & P             & R             & F             &                         \\ \hline
\begin{tabular}[c]{@{}c@{}}Jaderberg~\smallcite{jaderberg2016reading}\end{tabular}     & --          & --          & --          & --            & --            & --            & --            & --            & --            & --            & --            & --            & 7.3                      \\ \hline
\begin{tabular}[c]{@{}c@{}}MSERs-CNN~\smallcite{huang2014robust}\end{tabular}     & 0.88          & 0.71          & 0.78          & --            & --            & --            & --            & --            & --            & --            & --            & --            & --                      \\ \hline
\begin{tabular}[c]{@{}c@{}}MMser\\ \smallcite{zamberletti2014text}\end{tabular}         & --            & --            & --            & --            & --            & --            & 0.86          & 0.70          & 0.77          & --            & --            & --            & 0.75                    \\ \hline
\begin{tabular}[c]{@{}c@{}}TextFlow~\smallcite{tian2015text}\end{tabular}      & 0.86          & 0.76          & 0.81          & --            & --            & --            & 0.85          & 0.76          & 0.80          & --            & --            & --            & 1.4                     \\ \hline
\begin{tabular}[c]{@{}c@{}}FCRNall+filts\\ \smallcite{gupta2016synthetic}\end{tabular} & --            & --            & --            & \textbf{0.92} & 0.75          & 0.82          & --            & --            & --            & \textbf{0.92} & 0.76          & 0.83          & $>$1.27                   \\ \hline
\begin{tabular}[c]{@{}c@{}}FCN~\smallcite{Zhang_2016_CVPR}\end{tabular}         & --            & --            & --            & --            & --            & --            & 0.88          & 0.78          & 0.83          & --            & --            & --            & 2.1                     \\ \hline
\begin{tabular}[c]{@{}c@{}}SSD~\smallcite{liu2015ssd}\end{tabular}                                                     & --            & --            & --            & --            & --            & --            & 0.80          & 0.60          & 0.68          & 0.80          & 0.60          & 0.69          & 0.1                     \\ \hline
Fast TextBoxes                                                  & 0.86          & 0.74          & 0.80          & 0.88          & 0.74          & 0.80          & 0.86          & 0.74          & 0.80          & 0.88          & 0.74          & 0.81          & \textbf{0.09}           \\ \hline
TextBoxes                                                       & \textbf{0.88} & \textbf{0.82} & \textbf{0.85} & 0.89          & \textbf{0.82} & \textbf{0.86} & \textbf{0.88} & \textbf{0.83} & \textbf{0.85} & 0.89          & \textbf{0.83} & \textbf{0.86} & 0.73                    \\ \hline
\end{tabular}
\end{table*}

We replace the original TextBoxes detection score with the one in Eq.~\ref{eq:ctc-score}. However, evaluating Eq.~\ref{eq:ctc-score} on all boxes would be time-consuming. In practice, we first use TextBoxes to produce a redundant set of word candidates by detecting with a lower score threshold and a high NMS overlap threshold, preserving about 35 bounding boxes per image with a high recall of 0.93 with multi-scale inputs for ICDAR 2013 . Then we apply Eq.~\ref{eq:ctc-score} to all candidates to re-evaluate their scores, followed by a second score thresholding and a NMS. When dealing with multi-scale inputs, we generate candidates separately on each scale and perform the above steps on candidates of all the scales. 
Here we also adopt a slightly different NMS scheme. A lower overlap threshold is employed for boxes that are recognized as the same word, so that stronger suppression is imposed on boxes of the same word.

\begin{figure*}[!ht]
\begin{center}
\includegraphics[width=0.8\linewidth]{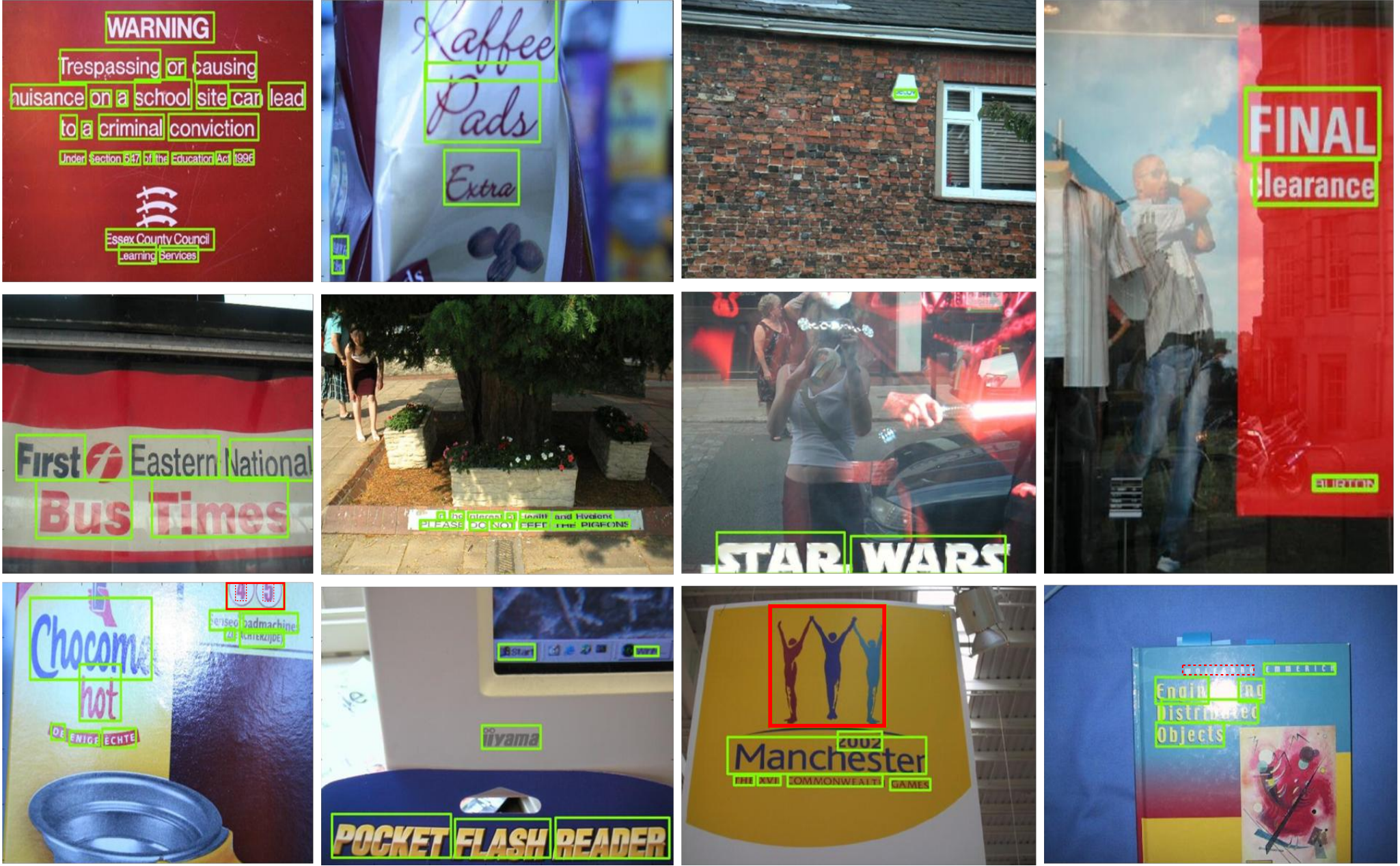}
\end{center}
\vspace{-5mm}
\caption{Examples of text localization results.
The green bounding boxes are correct detections; Red boxes are false positives; Red dashed boxes are false negatives.
}
\label{fig:localization-visualization}
\vspace{-2mm}
\end{figure*}

\section{Experiments}

We verify the effectiveness of TextBoxes on three different tasks, including text detection, word-spotting, and end-to-end recognition.

\subsection{Datasets}

\textbf{SynthText}~\cite{gupta2016synthetic} contains 800k synthesized text images, created via blending rendered words with natural images. The synthesized images look realistic, as the location and transform of text are carefully chosen with a learning algorithm. This dataset is used for pre-training our model.

\noindent\textbf{ICDAR 2011 (IC11)}~\cite{shahab2011icdar} There are real-world images with high resolution in the ICDAR 2011 dataset. The test set of the ICDAR 2011 dataset is used to evaluate our model.

\noindent\textbf{ICDAR 2013 (IC13)}~\cite{karatzas2013icdar}  The ICDAR 2013 dataset is similar to the ICDAR 2011 dataset. We use the training set of the ICDAR 2013 for training when we do experiments on the ICDAR 2011 dataset and the ICDAR 2013 dataset. The ICDAR 2013 dataset gives 3 lexicons of different sizes for the task of word spotting and end-to-end recognition. For each test image, it gives 100 words as a lexicon, which is called a strong lexicon. For the whole test set, it gives a lexicon containing hundreds of words, which is called a weakly lexicon. It also gives a generic lexicon which contains 90k words.

\noindent\textbf{Street View Text (SVT)}~\cite{wang2010word} The SVT dataset is more challenging than the ICDAR datasets due to the lower resolution of the images. There exist some unlabeled texts in the images. Thus, we only use this dataset for word spotting, in which a lexicon containing 50 words is provided for each image.


\subsection{Implementation details}
TextBoxes is trained with 300*300 images using stochastic gradient descent (SGD). Momentum and weight decay are set to $0.9$ and $5\times 10^{-4}$ respectively. Learning rate is initially set to $10^{-3}$, and decayed to $10^{-4}$ after 40k training iterations. On all the datasets except SVT, we first train TextBoxes on SynthText for 50k iterations, then finetune it on ICDAR 2013 training dataset for 2k iterations. On SVT, the finetuning is performed on the SVT training dataset. All training images are augmented online with random crop and flip, following the scheme in~\cite{liu2015ssd}.
All the experiments are carried out on a PC with one Titan X GPU. The whole training time is about 25 hours.
Text recognition is performed with a pre-trained CRNN~\cite{shi2015end} model\footnote{\url{https://github.com/bgshih/crnn}}, which is implemented and released by the authors. 

\begin{table*}[!tbp] \small
\centering
\caption{Word spotting and end-to-end results. The values in the table are F-measure. For ICDAR 2013, strong, weak and generic mean a small lexicon containing 100 words for each image, a lexicon containing all words in the whole test set and a large lexicon respectively. We use a lexicon containing 90k words as our generic lexicon. The methods marked by ``*'' are published on the ICDAR 2015 Robust Reading Competition website \url{http://rrc.cvc.uab.es}.}
\label{table:word spotting}
\renewcommand{\arraystretch}{1.2}
\begin{tabular}{|c|c|c|c|c|c|c|c|c|c|}
\hline
\multirow{2}{*}{Methods}                                                          & \multirow{2}{*}{\begin{tabular}[c]{@{}c@{}}IC11\\ spotting\end{tabular}} & \multirow{2}{*}{\begin{tabular}[c]{@{}c@{}}SVT\\ spotting\end{tabular}} & \multirow{2}{*}{\begin{tabular}[c]{@{}c@{}}SVT-50\\ spotting\end{tabular}} & \multicolumn{3}{c|}{\begin{tabular}[c]{@{}c@{}}IC13 \\ spotting\end{tabular}}& \multicolumn{3}{c|}{\begin{tabular}[c]{@{}c@{}}IC13 \\ end-to-end\end{tabular}} \\ \cline{5-10} 
                                                                                  &                                                                                     &                                                                              &                                                                                 & strong                                                            & weak        & generic       & strong                      & weak                        & generic                       \\ \hline
\begin{tabular}[c]{@{}c@{}}Alsharif~\smallcite{alsharif2013end}\end{tabular}         & --                                                                                  & --                                                                           & 0.48                                                                            & --                                                                  & --            & --            & --                            & --                            & --                            \\ \hline
\begin{tabular}[c]{@{}c@{}}Jaderberg~\smallcite{jaderberg2016reading}\end{tabular}   & 0.76                                                                                & 0.56                                                                         & 0.68                                                                            & --                                                                  & --            & 0.76          & --                            & --                            & --                            \\ \hline
\begin{tabular}[c]{@{}c@{}}FCRNall+filts\\ \smallcite{gupta2016synthetic}\end{tabular} & 0.84                                                                                & 0.53                                                                         & 0.76                                                                            & --                                                                  & --            & 0.85          & --                            & --                            & --                            \\ \hline
Deep2Text II+*  & --                                                                                & --                                                                        & --                                                                            & 0.85                                                                 & 0.83            & 0.80          & 0.82                            & 0.79                            & 0.77                            \\ \hline
SRC-B-TextProcessingLab*  & --                                                                                & --                                                                        & --                                                                            & 0.90                                                                 & 0.88            & 0.81          & 0.87                            & 0.85                            & 0.80                            \\ \hline
Adelaide\_ConvLSTMs*  & --                                                                                & --                                                                        & --                                                                            & 0.91                                                                 & 0.90            & 0.83          & 0.87                            & 0.86                            & 0.80                            \\ \hline
TextBoxes                                                                         & \textbf{0.87}                                                                       & \textbf{0.64}                                                                & \textbf{0.84}                                                                   & \textbf{0.94}                                                       & \textbf{0.92} & \textbf{0.87} & \textbf{0.91}                 & \textbf{0.89}                 & \textbf{0.84}                 \\ \hline
\end{tabular}
\end{table*}

\subsection{Text localization}
TextBoxes is tested on ICDAR 2011 and ICDAR 2013 for evaluating its text localization performance. The results are summarized and compared with other methods in Table.~\ref{tab:text-localization}. Results are evaluated under two different evaluation protocols, the DetEval~\cite{Wolf2006} and the ICDAR 2013 evaluation~\cite{karatzas2013icdar}.

Since there is a trade-off between precision and recall rate, f-measure is the most accurate measurement of detection performance. TextBoxes consistently outperforms competing methods in terms of f-measure. On ICDAR 2011, TextBoxes outperforms the second best methods~\cite{gupta2016synthetic}, by 4 percents. On ICDAR 2013, TextBoxes also outperforms competing methods by at least 2 percents. TextBoxes ranks the first in term of testing speed, even with the multi-scale version, which takes only 0.73s per image. Meanwhile, a fast implementation of TextBoxes takes merely 0.09s per image, without much loss in accuracy.

In order to further verify the effectiveness of TextBoxes, we also report the results of SSD~\cite{liu2015ssd} for the comparison in Table.~\ref{tab:text-localization}, which is the most relevant and the state-of-the-art detector for general objects.   
Here, SSD is trained using the same procedures as TextBoxes. SSD achieves competitive performance, but still falls short of other state-of-the-art methods. In particular, we observe that SSD cannot achieve good results when detecting words with large aspect ratios while TextBoxes performs much better, benefiting from the proposed text-box layers which are designed in order to overcome the length variation of words.  　　

\begin{figure*}[!ht]
\begin{center}
\includegraphics[width=0.8\linewidth]{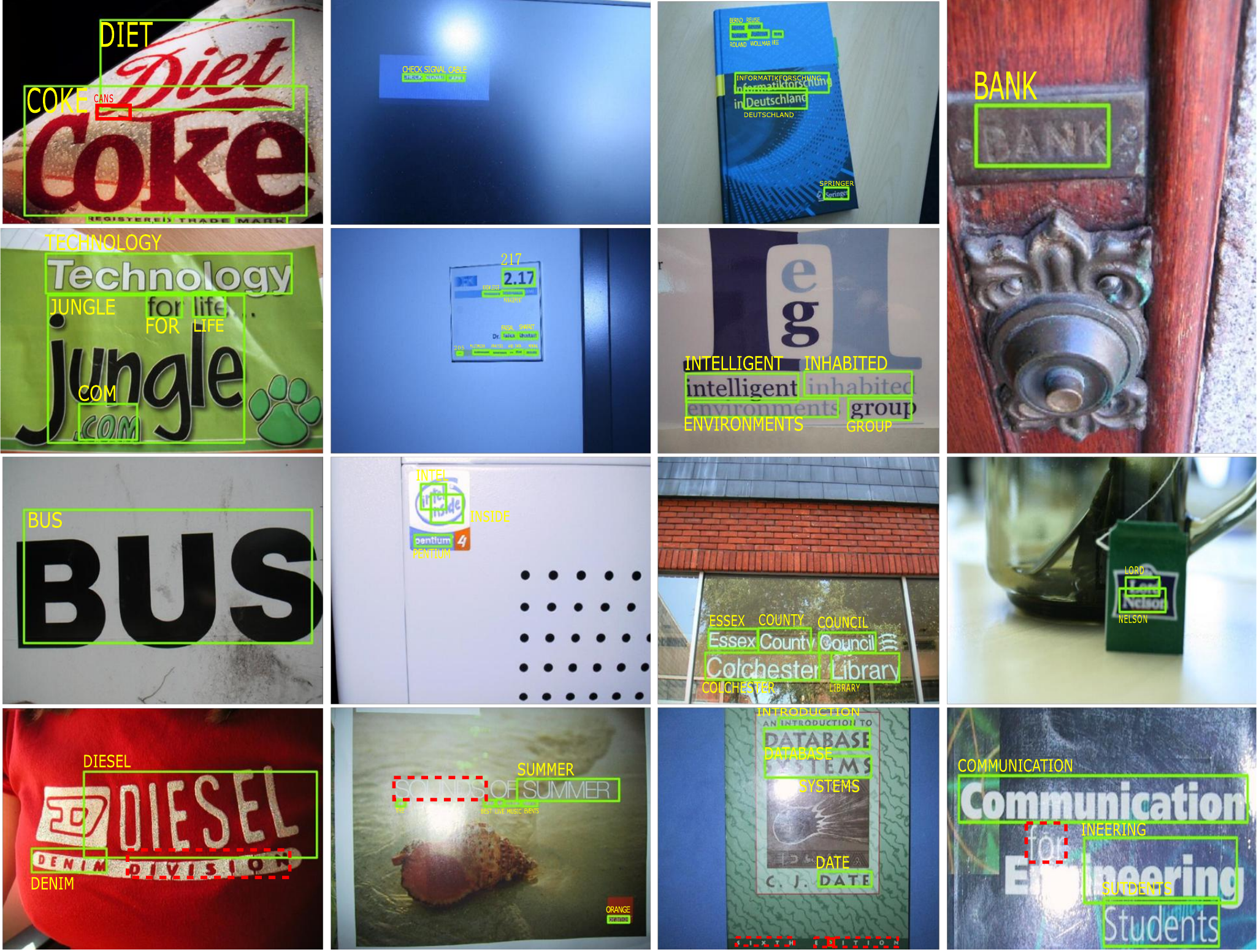}
\end{center}
\vspace{-5mm}
\caption{Examples of word spotting results. Yellow words are recognition results. Words less than 3 letters are ignored, following the evaluation protocol.
The box colors have the same meaning as Fig.~\ref{fig:localization-visualization}.}
\label{fig:spotting-visualization}
\vspace{-2mm}
\end{figure*}

\subsection{Word spotting and end-to-end recognition}
The performance of word spotting is evaluated by detection results that are refined by recognition, while the evaluation of end-to-end performance concerns both detection and recognition results. We test TextBoxes on ICDAR 2011, SVT, and ICDAR 2013.

As shown in Table.~\ref{table:word spotting}, our method outperforms all the existing methods, including the most recent competition results published on the website.
On ICDAR 2011 and ICDAR 2013, our method outperforms the second best method at least 2 percents with all the evaluation protocol listed in Table.~\ref{table:word spotting}.
The performance gap on SVT is even larger. TextBoxes outperforms the leading method~\cite{gupta2016synthetic}, by over 8 percents on both SVT and SVT-50.
The reason is likely to be that TextBoxes is more robust to the low resolution images in SVT, since TextBoxes is trained on relatively low resolution images.

Coupled with a recognition model, TextBoxes achieves the state-of-the-art performance on end-to-end recognition benchmarks. On ICDAR 2013, TextBoxes breaks the records recently made by Adelaide\_ConvLSTMs* on all the lexicon settings. More specifically, TextBoxes generates about 35 proposals per image when using multi-scale inputs on ICDAR 2013, with a recall of 0.93. With a strong lexicon for the recognition model, 3.8 bounding boxes per image are reserved, achieving a recall of 0.91 and a precision of 0.97. 
We employ a 90k-lexicon for SVT and ICDAR 2011, and a 50-word lexicon per image on SVT-50. Note that even though Jaderberg~\cite{jaderberg2016reading} and FCRNall+filts~\cite{gupta2016synthetic} adopt a much smaller lexicon(50k words), their results are still inferior to our method.

\subsection{Running speed}
Most existing methods detect texts in a multi-step manner, making them hard to run efficiently. Most of the computation of TextBoxes is spent on the convolutional forward passes, which are very fast when running on GPU devices.
TextBoxes takes only 0.09s per image with $700*700$ single-scale images, resulting in an f-measure of 0.80 on ICDAR 2013, which is still very competitive.
When running on 5 input scales, TextBoxes achieves 0.85 f-measure on ICDAR 2013, taking 0.73 second per image with the batch size setting to 1.
We remove the $1600*1600$ scale when testing on SVT, since the SVT image resolutions are relatively low.
Testing on the the remaining scales takes merely 0.24 second per image.

The speed comparisons are listed in Table.~\ref{tab:text-localization}. \cite{jaderberg2016reading} adopts two proposal generation methods, a random forest classifier, and a CNN regression model. They each takes 1-3 seconds, about 7s in total.
\cite{gupta2016synthetic} proposes a YOLO-like model called FCRN, followed by the same random forest classifiers and a CNN regression model. It takes 1.27s excluding the regression step, whose running time is not reported.
TextBoxes achieves the highest detection accuracy while being the fastest among them.

\subsection{Weaknesses}
TextBoxes performs well in most situations. However, it still fails to handle some difficult cases, such as overexposure and large character spacing. Some failure cases are shown in Fig.~\ref{fig:localization-visualization} and Fig.~\ref{fig:spotting-visualization}.

\section{Conclusion}
We have presented TextBoxes, an end-to-end fully convolutional network for text detection, which is highly stable and efficient to generate word proposals against cluttered backgrounds. Comprehensive evaluations and comparisons on benchmark datasets clearly validate the advantages of Textboxes
in three related tasks including text detection, word spoting and end-to-end recognition. In the future, we are interested to extend TextBoxes for multi-oriented texts, and combine the networks of detection and recognition into one unified framework.


\section{Acknowledgements}
This work was partly supported by National Natural Science Foundation of China (61222308, 61573160, 61572207 and 61503145), and Open Project Program of the State Key Laboratory of Digital Publishing Technology (F2016001).
\bibliographystyle{aaai}
\bibliography{TextBoxes}

\end{document}